\documentclass[runningheads, envcountsame, a4paper]{llncs}
\usepackage{graphicx}
\PassOptionsToPackage{hyphens}{url}\usepackage{hyperref}
\usepackage{xurl}
\usepackage{epstopdf}
\usepackage{microtype}

\usepackage{subcaption}
\usepackage{multirow}
\usepackage{amsmath}
\usepackage{amsfonts}
\makeatletter
\newcommand{\printfnsymbol}[1]{%
  \textsuperscript{\@fnsymbol{#1}}%
}
\makeatother

\begin{document}
\title{Conditioned Text Generation with Transfer for Closed-Domain Dialogue Systems}
\titlerunning{Conditioned Text Generation with Transfer}
%
\author{St\'{e}phane d'Ascoli\inst{1,2}\thanks{Both authors contributed equally.} \and
Alice Coucke\inst{1}\printfnsymbol{1} \and
Francesco Caltagirone\inst{1} \\ \and
Alexandre Caulier\inst{1} \and Marc Lelarge\inst{2,3}}
\authorrunning{S. d'Ascoli et al.}
%
\institute{Sonos Inc., Paris, France \email{\{alice.coucke,francesco.caltagirone,alexandre.caulier\}@sonos.com}
\and
ENS, CNRS, PSL University, Paris, France \email{stephane.dascoli@ens.fr}
 \and
INRIA, Paris, France \email{marc.lelarge@ens.fr}}
\maketitle              

\begin{abstract}
    Scarcity of training data for task-oriented dialogue systems is a well known problem that is usually tackled with costly and time-consuming manual data annotation. An alternative solution is to rely on automatic text generation which, although less accurate than human supervision, has the advantage of being cheap and fast.
    Our contribution is twofold. First we show how to optimally train and control the generation of intent-specific sentences using a conditional variational autoencoder. Then we introduce a new protocol called \textit{query transfer} that allows to leverage a large unlabelled dataset, possibly containing irrelevant queries, to extract relevant information. Comparison with two different baselines shows that this method, in the appropriate regime, consistently improves the diversity of the generated queries without compromising their quality. We also demonstrate the effectiveness of our generation method as a data augmentation technique for language modelling tasks.

\end{abstract}

\section{Introduction}
    Closed-domain dialogue systems have become ubiquitous nowadays with the rise of conversational interfaces. These systems aim at extracting relevant information from a user's spoken query, produce the appropriate response/action and, when applicable, start a new dialogue turn. The typical spoken language understanding (SLU) framework relies on a speech-recognition engine that transforms the spoken utterance into text followed by a natural language understanding engine that extracts meaning from the text utterance. 
    
    Here we consider essentially single-turn closed-domain dialogue systems where the meaning is well summarized by an intent and its corresponding slots. As an example, the query ``Play Skinny Love by Bon Iver'' should be interpreted as an intent {\it PlayTrack} with slots {\it TrackTitle} ``Skinny Love'' and {\it Artist} ``Bon Iver''.
    
    Training data for conversational systems consists in utterances together with their annotated intents. In order to develop a new interaction scheme with new intents, a (possibly large) representative set of manually annotated utterances needs to be produced, which is a costly and time-consuming process. It is therefore desirable to automate it as much as possible. We consider the scenario in which only a small set of annotated queries is available for all the in-domain intents, but we also have access to a large ``reservoir'' dataset of unannotated queries that belong to a broad spectrum of intents ranging from close to far domain. This situation is indeed very typical of conversational platforms like Google's DialogFlow or IBM Watson which offer a high degree of user customization.

    \subsection{Contribution and Outline}
    
        We focus on automatic generation of utterances conditioned to desired intents, aiming to alleviate the problem of training data scarcity. Using a Conditional Variational Auto-Encoder~\cite{cvae_2015} (CVAE), we show how it is possible to selectively extract the valuable information from the reservoir dataset. We call this mechanism {\it query transfer}. We analyse the performance of this approach on the publicly-available Snips benchmark dataset~\cite{coucke2018snips} through both quality and diversity metrics. We also observe an improvement in the perplexity of a language model trained on data augmented with our generation scheme. This preliminary result is encouraging for future application to SLU data augmentation. 
            
        The paper is structured as follows: in Section \ref{sec: rel_work}, we briefly present the related literature, in Section \ref{sec: approach} we introduce our approach, and in Section \ref{sec: results} we show our experimental results before concluding in Section \ref{sec: conclusions}. 

    \subsection{Related work}
        \label{sec: rel_work}
        
        While there is a vast literature on conditional text generation, semi-supervised learning and data augmentation, there are only few existing works that combine these elements.  Shortly after the Variational Autoencoder (VAE) model was introduced by~\cite{kingma2013auto}, a conditional variation autoencoder (CVAE) model was used for semi-supervised classification tasks by~\cite{kingma2014semi} and later improved by \cite{jang2016categorical}.
        
        \cite{kurata2016,hou2018sequence,hu2017toward} generate utterances through paraphrasing with the objective of augmenting the training set and improving slot-filling or other NLP tasks without conditioning on the intent. The data used to train the paraphrasing model is annotated and in-domain.
        \cite{yoo2019slot} leverage a CVAE to generate queries conditioned to the presence of certain slots and observe improvements in slot-filling performance when augmenting the training set with generated data. \cite{yoo2019slu} propose instead an AE that is capable of jointly generating a query together with its annotation (intent and slots) and show improvements in  intent classification and slot-filling through data augmentation.
        
        In a recent paper, semi-supervised self-learning is used to iteratively incorporate data from an unannotated set into the annotated training set~\cite{cho2019}. Their chosen metrics are both SLU performance and query diversity. This method represents a valid alternative to our protocol and will be the object of competitive benchmarks in future work.

\begin{figure*}[htb]
        \centering
        \begin{subfigure}{0.55\linewidth}
            \includegraphics[width=1\textwidth]{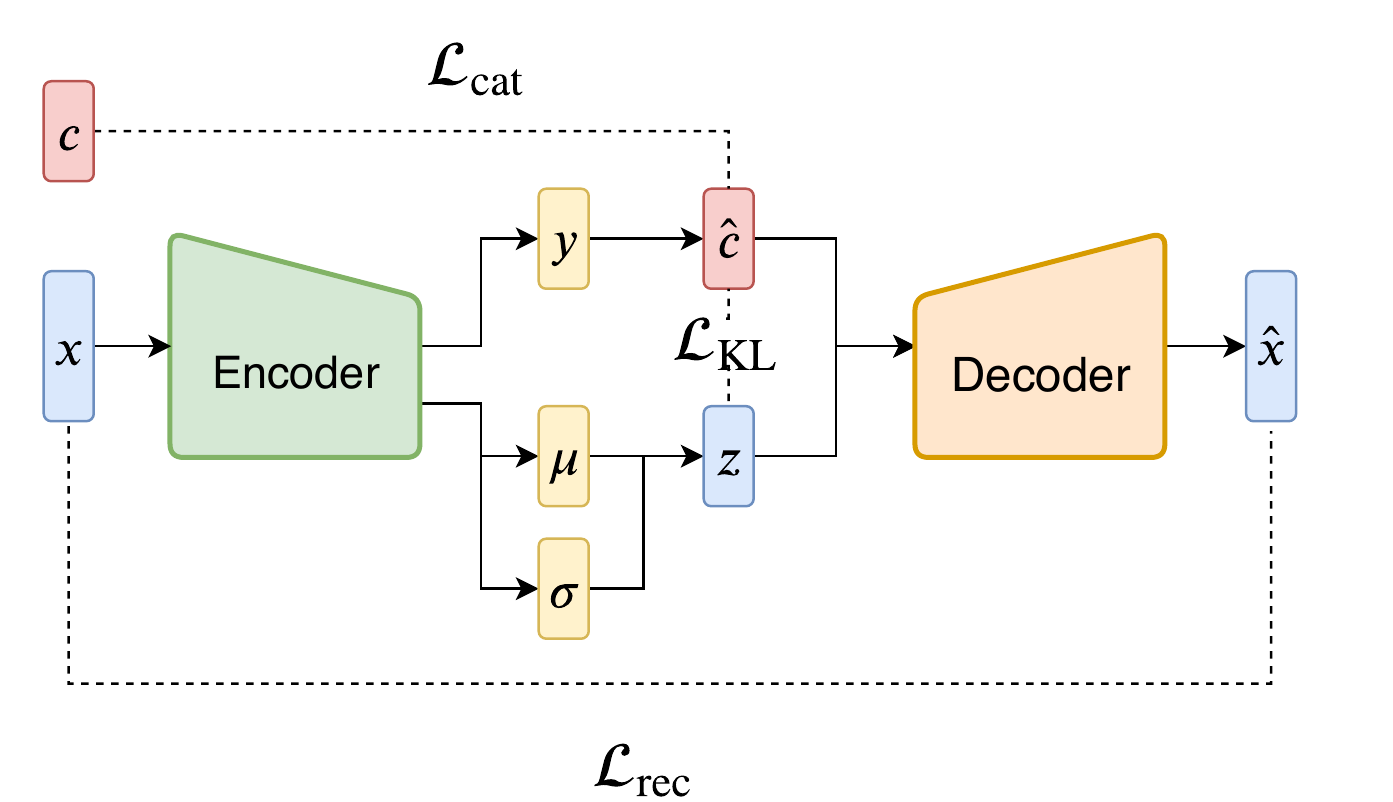}
        \end{subfigure}
        \hfill
        \begin{subfigure}{0.4\linewidth}
            \includegraphics[width=1\textwidth]{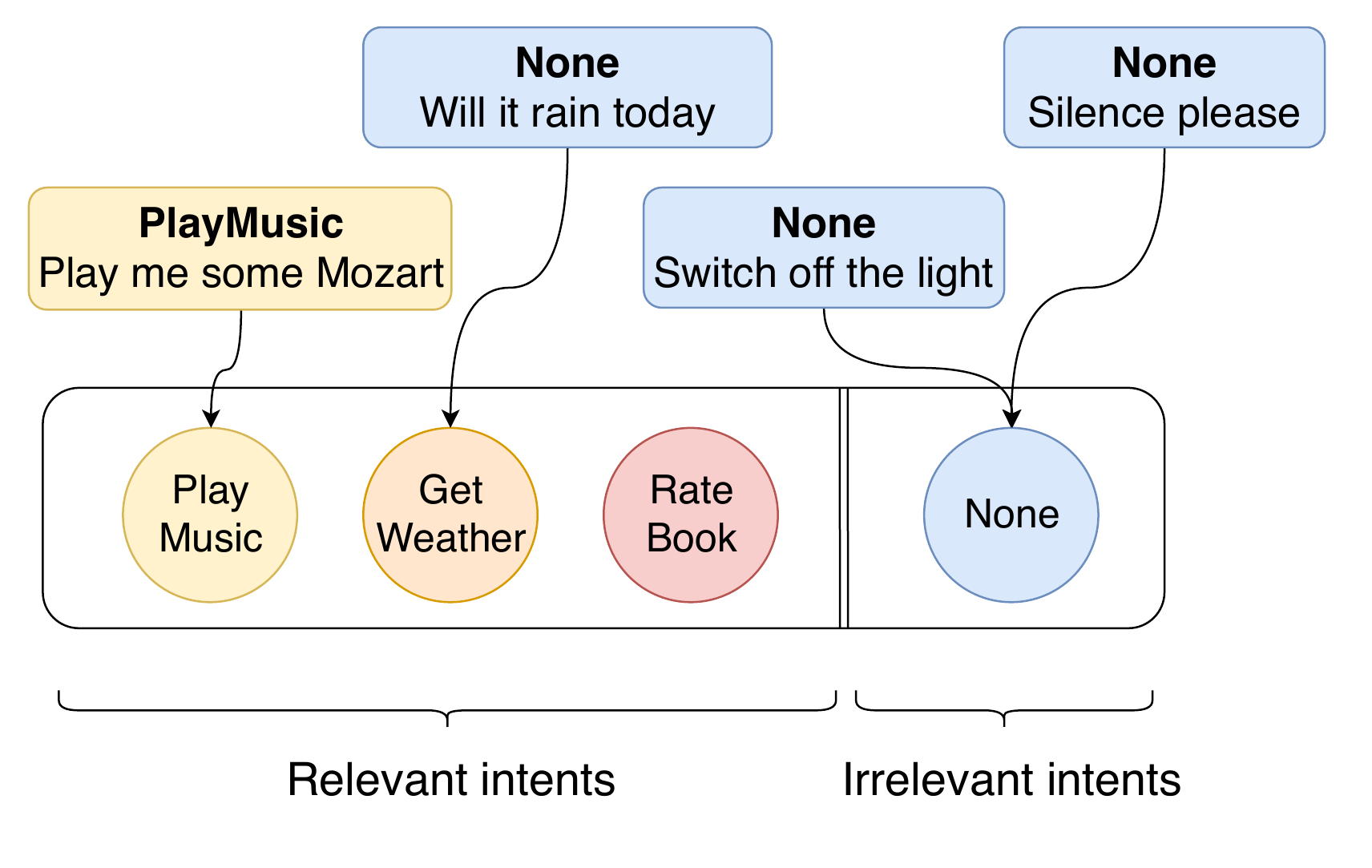}
        \end{subfigure}
        \caption{Architecture of the model. (Left panel) The variational autoencoder architecture with the various losses defined in Section~(\ref{sec: approach}). 
        (Right panel) An illustration of the categorical latent vector filtering relevant sentences}
        \label{fig:model}
    \end{figure*}

\section{Approach}
    \label{sec: approach}
    
    \subsection{Conditional variational autoencoders} 
        In order to generate queries conditioned to an underlying intent, we use a CVAE as depicted in~Fig.~\ref{fig:model} (left), with a continuous code $\boldsymbol{z}$ and a discrete code $\boldsymbol{c}$ thanks to the Gumbel softmax trick~\cite{maddison2016gumbel,jang2016categorical}. Each training sample consists in a continuous feature vector $\boldsymbol{x}$ and a categorical label expressed as a one-hot vector $\boldsymbol{c}$. The latent code is the concatenation of a continuous latent vector $\boldsymbol{z}$ and a predicted categorical vector $\boldsymbol{\hat{c}}$.
    	The associated loss function consists of three terms: the reconstruction term, the regularization  term, and the supervision term for the label:
    	\begin{align*}
    	\mathcal{L} &= \mathcal{L}_{\rm rec} + \gamma \mathcal{L}_{\rm KL} + \mathcal{L}_{\rm cat},
    	\\
    	    \mathcal{L}_{\rm rec} &= -\underset{q_{\phi}(\boldsymbol{z}| \boldsymbol{x})}{\mathbb{E}}[\log p_{\theta}(\boldsymbol{x} | \boldsymbol{z}, \boldsymbol{\hat{c}})], \\
    	    \mathcal{L}_{\rm KL} &= D_{\mathrm{KL}}(q_{\phi}(\boldsymbol{z} | \boldsymbol{x}) \| p(\boldsymbol{z})) +D_{\mathrm{KL}}(q_{\phi}(\boldsymbol{\hat{c}} | \boldsymbol{x}) \| p(\boldsymbol{\hat{c}})),\\
    	    \mathcal{L}_{\rm cat} &= -\sum_{i=1}^{C} \hat{c}_i \, \alpha_{i} \,\log \left(q_{\phi}(\hat{c}_i | \boldsymbol{x}) \right).
    	\label{eq:losses}
    	\end{align*}
    	
    	$q_{\phi}$ and $p_{\theta}$ represent the encoder and the decoder with their associated parameters, and $C$ is the dimension of the categorical latent space.
    	The constant $\gamma$ is used to set the relative weight of the KL regularization and perform annealing during training \cite{bowman2015generating} \cite{sonderby2016train}. The class-specific $\alpha$ coefficients will play a crucial role for the \textit{query transfer} mechanism described below. 
    	In all our experiments, $p(\boldsymbol{\hat{c}})$ is uniform and $p(\boldsymbol{z}) = \mathcal{N}(\vec 0,1)$. At inference time, a sentence is generated by feeding the decoder with the concatenation of a chosen $\boldsymbol{\hat{c}}$ with a sampled $\boldsymbol{z}$ and extracting greedily the most probable sequence.
	
	\subsection{Query transfer}
	    
	    \subsubsection{Approach.}
    	A CVAE can be trained on a dataset of annotated queries $(\boldsymbol{x}, \boldsymbol{c})$, where $\boldsymbol{x}$ is the sentence and $\boldsymbol{c}$ is the underlying query's intent.
    	With too few sentences for training, a CVAE will not yield generated sentences of high enough quality and diversity. In addition to an annotated training dataset $\mathcal{D}_0$ -- kept small in the data scarcity regime of interest in this paper -- 
    	a large ``reservoir'' dataset $\mathcal{D}_r$ is considered. The latter is unannotated and contains sentences that potentially cover a larger spectrum, ranging from examples that are semantically close to the in-domain ones to completely out-of-domain examples. 
        The categorical latent space of the CVAE contains one dimension for each intent in $\mathcal{D}_0$. The novelty in our approach is to allocate an extra dimension for irrelevant sentences coming from $\mathcal{D}_r$, namely an additional \textit{None} intent. All sentences from $\mathcal{D}_r$ are supervised to this dimension by the cross-entropy loss, but we want the relevant ones to be allowed to transfer to one of the intents of $\mathcal{D}_0$, as illustrated in~Fig.~\ref{fig:model} (right). To control this, we adjust the amount of transfer by multiplying the supervision loss of $\mathcal{D}_r$ by a factor $\alpha$ (we will take $\alpha_{\mathrm{None}}=\alpha$ and $\alpha_{i\neq None} = 1 $). In the case $\alpha = 0$, the sentences from $\mathcal{D}_r$ are not supervised at all. 
    
        \subsubsection{Illustration on images.}
        \label{sec:illu_images}
        The transfer process and the effect of $\alpha$ are illustrated here in the context of computer vision. We present results on the MNIST and Fashion MNIST~\cite{xiao2017fashion} datasets as toy examples. Here, the small annotated dataset $\mathcal{D}_0$ contains only examples from the first 6 classes of each dataset, with 10 examples per class ($0-5$ digits and 6 items of clothing). The larger reservoir dataset $\mathcal{D}_r$ contains examples from each of the 10 classes (half of its content being therefore irrelevant to the generative task), with 50 examples per class.

    	\begin{figure}[htb]
    	    \centering
    		\begin{subfigure}[b]{0.23\textwidth}
    			\includegraphics[width=\textwidth]{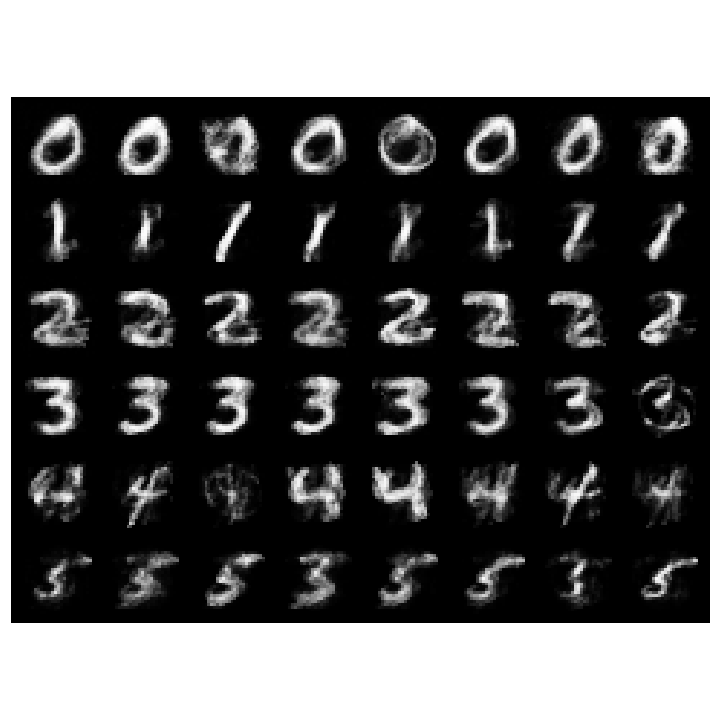}
    			\caption{Without $\mathcal{D}_r$}
    		\end{subfigure}
    		\begin{subfigure}[b]{0.23\textwidth}
    			\includegraphics[width=\textwidth]{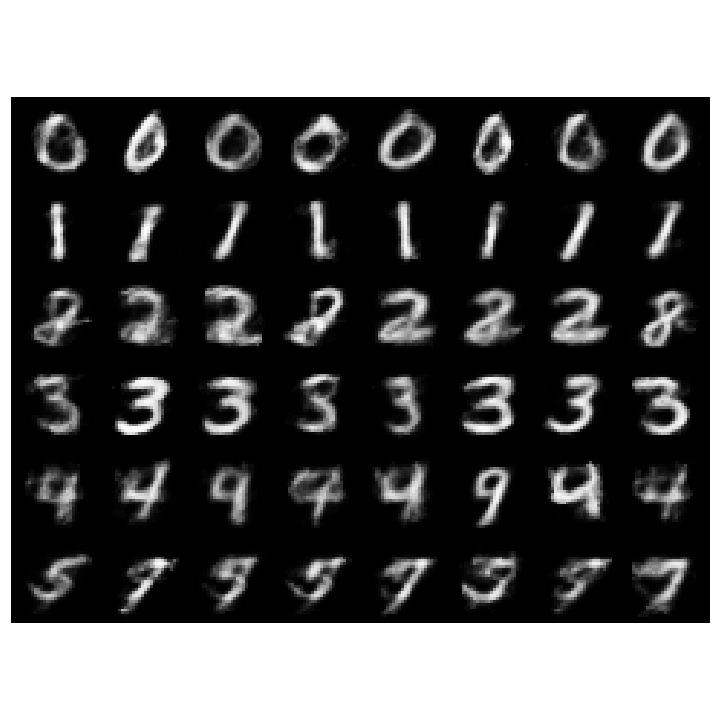}
    			\caption{$\mathcal{D}_r$, $\alpha=0$}
    		\end{subfigure}
    		\begin{subfigure}[b]{0.23\textwidth}
    			\includegraphics[width=\textwidth]{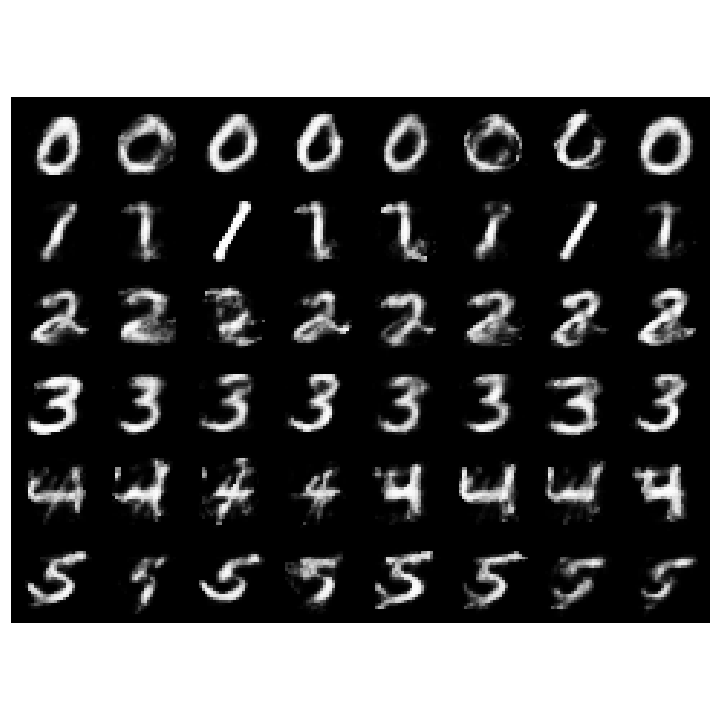}
    			\caption{$\mathcal{D}_r$, $\alpha=2$}
    		\end{subfigure}
    		\begin{subfigure}[b]{0.23\textwidth}
    			\includegraphics[width=\textwidth]{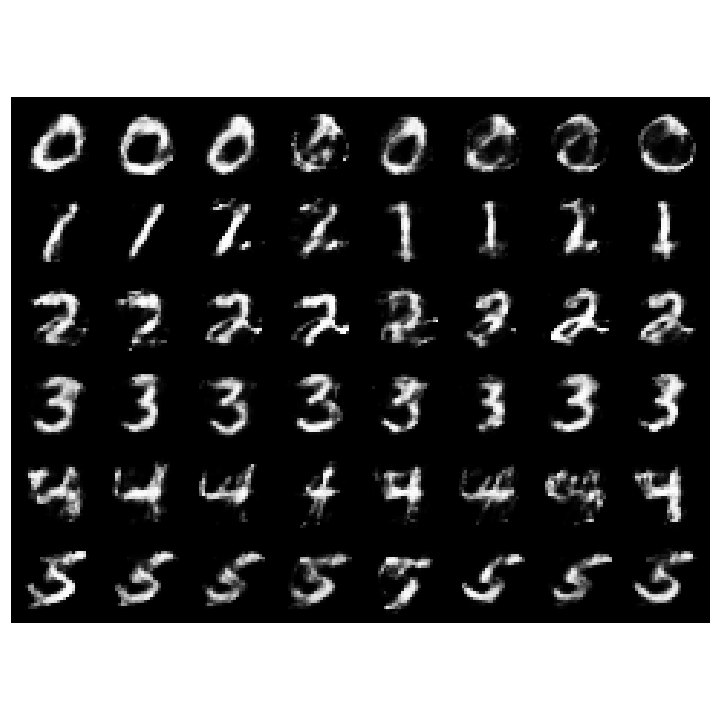}
    			\caption{$\mathcal{D}_r$, $\alpha=5$}
    		\end{subfigure}
            \begin{subfigure}[b]{0.23\textwidth}
    			\includegraphics[width=\textwidth]{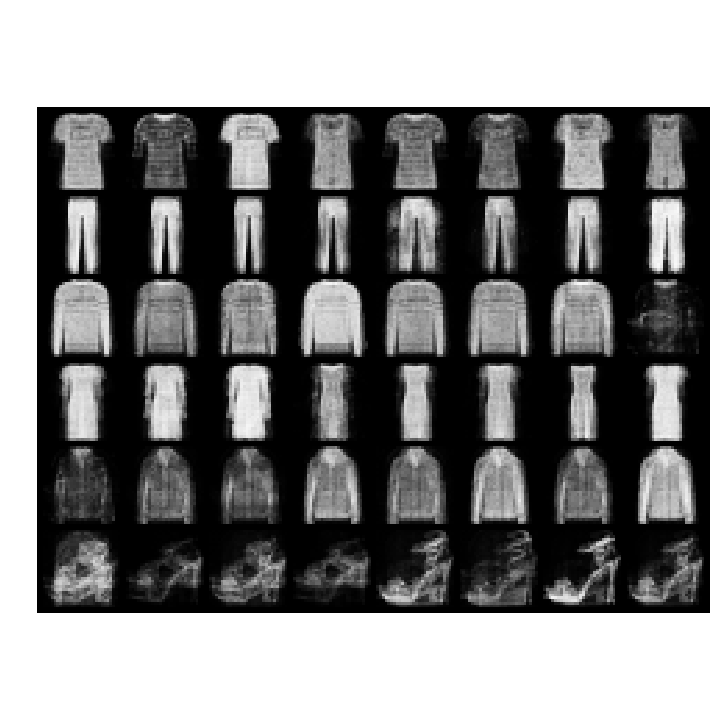}
    			\caption{Without $\mathcal{D}_r$}
    		\end{subfigure}
    		\begin{subfigure}[b]{0.23\textwidth}
    			\includegraphics[width=\textwidth]{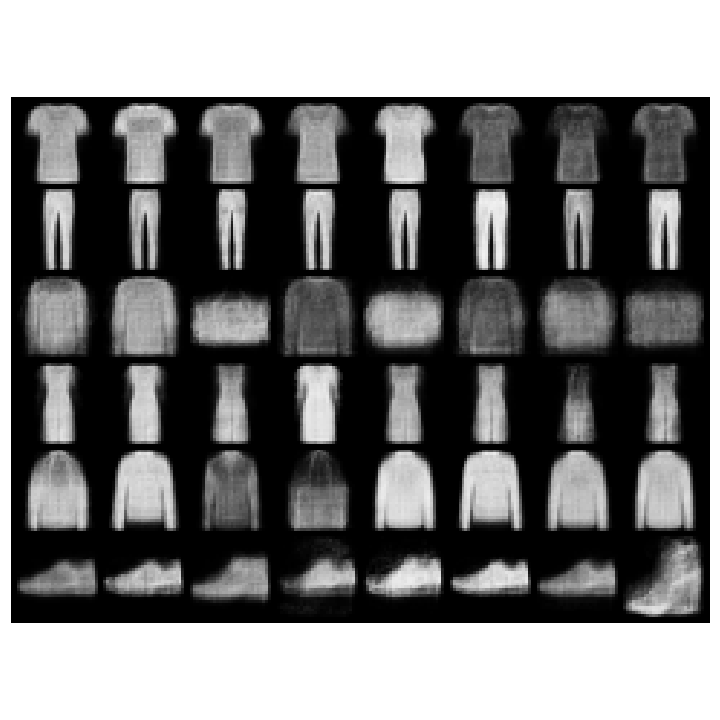}
    			\caption{$\mathcal{D}_r$, $\alpha=0$}
    		\end{subfigure}
    		\begin{subfigure}[b]{0.23\textwidth}
    			\includegraphics[width=\textwidth]{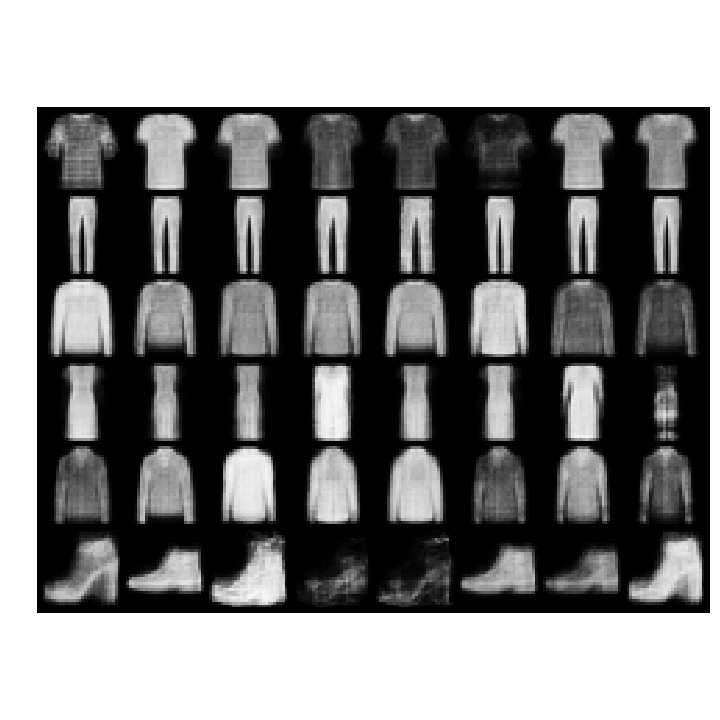}
    			\caption{$\mathcal{D}_r$, $\alpha=0.1$}
    		\end{subfigure}
    		\begin{subfigure}[b]{0.23\textwidth}
    			\includegraphics[width=\textwidth]{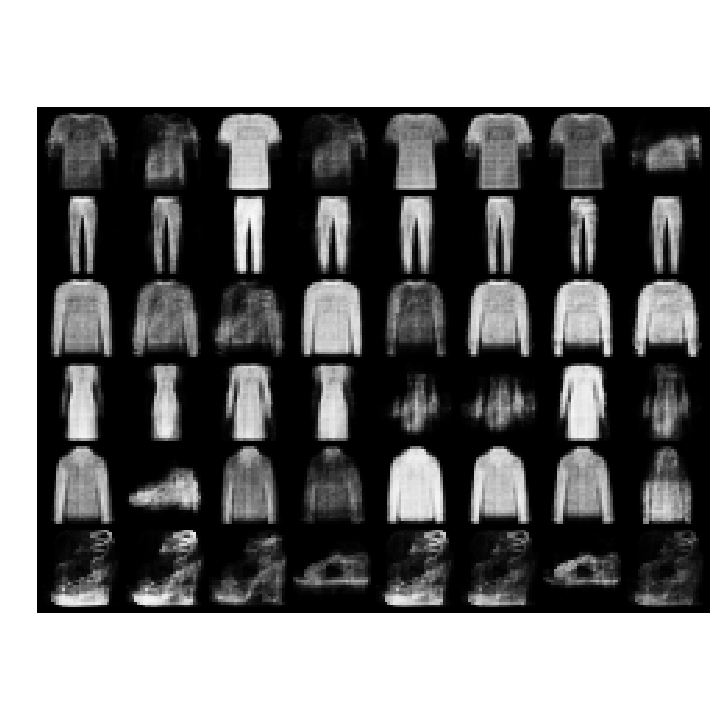}
    			\caption{$\mathcal{D}_r$, $\alpha=1$}
    		\end{subfigure}

    		\caption{Top row: MNIST dataset. Bottom row: Fashion MNIST dataset. In the leftmost panels, the CVAE is trained on a small labelled dataset $\mathcal{D}_0$ containing only the first 6 classes (10 images per class). In the other panels, the CVAE leverages an unlabelled reservoir dataset $\mathcal{D}_r$ containing all 10 classes (50 images per class), with a varying transfer parameter $\alpha$. The best quality-diversity trade-off is reached at $\alpha = 2$ for MNIST, and $\alpha=0.1$ for Fashion MNIST}
    	\label{fig:images}
    	\end{figure}
    	
    	Fig. \ref{fig:images} shows generated images obtained by training a very simple two-layer fully-connected CVAE for 200 epochs, for various values of the transfer parameter $\alpha$. 
    	We see that without the reservoir dataset $\mathcal{D}_r$ (panel (a) on both figures), there is not enough training data to generate high-quality diverse images. 
    	
    	Using $\mathcal{D}_r$ with a too low value of $\alpha$ (second column) yields unwanted image transfer and corruption of the generated images (unwanted 9's corrupt the 4's in MNIST, unwanted bags corrupt the shirts in Fashion MNIST). Conversely, if $\alpha$ is too high (panels (d) on both figures), there is not enough transfer and the generated images do not benefit from $\mathcal{D}_r$ anymore. However, for a well-chosen value $\alpha^\star$ (panels (c)), there is significant improvement both in quality and diversity of the generated images. As we can see here, the optimal value of $\alpha$ is dataset-dependent -- $\alpha^\star\sim2$ for MNIST, $\alpha^\star\sim 0.1$ for Fashion MNIST -- and needs to be tuned accordingly.

	\subsection{Sentence selection}\label{sec:sent_selec}
    
        To further improve the query transfer, we also introduce a sentence selection procedure in order to remove from $\mathcal{D}_r$ irrelevant data that can potentially pollute generation. In the context of natural language processing, this may be achieved by sentence embeddings. We use generalist sentence embeddings such as InferSent \cite{conneau2017supervised} as a first, rough, sentence selection mechanism. We first compute an ``intent embedding'' $\Vec{I}$ for each intent of $\mathcal{D}_0$, obtained by averaging the embeddings of all the sentences of the given intent. Then we only collect the sentences from $\mathcal{D}_r$ which are ``close'' enough to one of the intents of $\mathcal{D}_0$, i.e
            $\left\lbrace \Vec{S} \mid \exists \Vec{I}, \cos \left(\theta(\Vec{I}, \Vec{S})\right) > \beta \right\rbrace$,
        where $\beta$ is a threshold which controls selectivity.

\section{Experimental results}
    \label{sec: results}
    
    \subsection{Experimental setup}
        A PyTorch implementation of the CVAE model is publicly available on GitHub at \url{https://github.com/snipsco/automatic-data-generation}.
        
        \subsubsection{Datasets.}
            We use the publicly-available Snips benchmark dataset \cite{coucke2018snips}, which contains user queries from 7 intents such as \textit{PlayMusic} or \textit{GetWeather}, with 2000 queries per intent in the training set (from which we will only keep small fractions for $\mathcal{D}_0$ to mimic scarcity) and 100 queries per intent in the test set. Each intent comes with specific slots. 
            
            As a proxy for a reservoir dataset $\mathcal{D}_r$, we use a large in-house (private) dataset which contains all sorts of queries from nearly 350 varied English intents, typical of voice assistants. Each intent comes with at least 100 examples queries. Some intents may be more or less close semantically (e.g. \textit{RadioOn} and \textit{MusicOn} as opposed to \textit{SetOven}) or even duplicates with different intent names, but the only available label for $\mathcal{D}_r$ is the intent a given query belongs to.

        \subsubsection{Hyperparameters.}
        \label{sec:training}
            The GloVe embedding size is set to 100.
            Both the encoder and the decoder of our model use one-layer GRUs, with a hidden layer of size $256$ and both the continuous and categorical latent spaces are of size $8$ (for the categorical one: seven intents + one \textit{None} class).  At each step of the decoding sequence, a softmax with temperature $\tau=1$ is applied to the output of the decoder. We adopt the KL-annealing trick from \cite{bowman2015generating} to avoid posterior collapse: the weight of the KL loss term is annealed from $0$ to $1$ using the logistic function, at a time and a rate given by two hyper-parameters $t_{KL}$ and $r_{KL}$. The hyper-parameters were chosen to ensure satisfactory intent conditioning: $t_{KL}=300$ and $r_{KL}=0.01$, but were not optimized in any particular way since model selection is not particularly meaningful outside of a specific task. 
            
            The {\it Adam} optimizer is used and we train for $50$ epochs at a learning rate of $0.01$ with a batch size of $128$. Depending on the size of $\mathcal{D}_0$, it takes a few dozens of minutes per experiment on a laptop. No word or embedding dropout is applied. Note that we draw a fixed number of samples from both $\mathcal{D}_0$ and $\mathcal{D}_r$, however since we are in a data scarcity regime and only consider small $\mathcal{D}_0$, this draw entails high variability. Hence all results presented are averaged over five random seeds. For all of the experiments described in Section \ref{sec:results}, the size of the dataset is set to $|\mathcal{D}_0|=200$ and the sentence selection threshold to $\beta=0.9$. The $\alpha$ parameter allows to control the amount of transfer between $\mathcal{D}_0$ and the reservoir $\mathcal{D}_r$. 
            
        \subsubsection{Delexicalization procedure.}
            The word embeddings fed to the encoder are pre-trained GloVe embeddings~\cite{pennington2014glove}. We use a delexicalization procedure similar to that used in ~\cite{hou2018sequence} for Seq2Seq models.  First, slot values are replaced a placeholder and stored in a dictionary (``Weather in Paris'' $\rightarrow$ ``Weather in [City]''). The model is then trained on these delexicalized sentences and new delexicalized sentences are generated. The last step is to relexicalize the generated sentences: abstract slot names are replaced by stored slot values. The effort is indeed put on generating new contexts, rather than just shuffling slot values.
            
            We tried various strategies for the initialization of slot-embeddings (e.g. the average of all slot values) and found that it had no impact in our experiments. We therefore initialize them with random embeddings.
    	
    \subsection{Generation metrics}
        Our generation task must optimize a trade-off between quality and diversity: the generated sentences need to be consistent with the original dataset and to bring novelty. To account for quality, we first consider the \textbf{intent conditioning} accuracy. The generated sentences need to be well-conditioned to the intent imposed. 
        We train an intent classifier 
        on the full Snips training set, which reaches near-perfect accuracy on the test set, and use it as an ``oracle'' to evaluate intent conditioning accuracy, independently of the the generation mechanism.
        We then assess the semantic quality of the generated sentences by considering 
        the \textbf{BLEU-quality}, namely the forward Perplexity \cite{zhao2017adversarially}, or the BLEU score \cite{papineni2002bleu} computed against the reference sentences of the given intent.
        
        
        To account for diversity, we consider the \textbf{BLEU-diversity} defined as $1-\textrm{self-BLEU}$ where $\textrm{self-BLEU}$ is merely the BLEU score of the generated sentences of a given intent computed against the other generated sentences of the same intent \cite{zhu2018texygen}. However, enforcing high BLEU-diversity does not ensure that we are not just reproducing the training examples. Therefore, we also consider \textbf{originality}, defined as the fraction of generated sentences that are not present in the training set under their delexicalized form.  

        These four metrics take values in $\left[0 , 1\right]$. The three last metrics are evaluated intent-wise, which may be problematic if the intent conditioning of the generated sentences is poor. For example, if we condition to \textit{PlayMusic} and the generated sentence is ``What is the weather~?'', the diversity metrics of the \textit{PlayMusic} intent would be over-estimated while the quality would be under-estimated. To reduce this effect as much as possible, the computation of these metrics is therefore restricted to generated sentences for which the oracle classifier (used for the intent conditioning accuracy metric) agrees with the conditioning intent. 
         
    \subsection{Generation results}
        \label{sec:results}
        
        \subsubsection{Quality-diversity trade-off}

\begin{figure*}[htb]
            \centering
            \includegraphics[width=.49\textwidth]{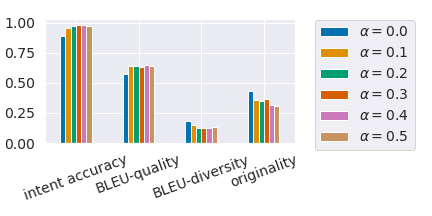}
            \includegraphics[width=.49\textwidth]{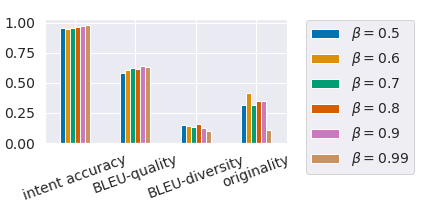}
            \caption{(Left Panel) Evolution of the generation metrics as a function of the transfer parameter $\alpha$ for $|\mathcal{D}_0| = |\mathcal{D}_r|$ and $\beta=0.9$. (Right Panel) Evolution of the generation metrics as a function of the transfer parameter $\beta$ for $|\mathcal{D}_0| = |\mathcal{D}_r|$ and $\alpha=0.2$}
            \label{fig:cursor}
        \end{figure*}

        Fig.~\ref{fig:cursor} shows that $\alpha$ (left panel) and $\beta$ (right panel) are useful cursors to control the diversty-quality tradeoff. Increasing them yields generated sentences of higher quality (both in terms of intent conditioning and BLEU-quality) but lower diversity (in terms of BLEU-diversity and originality). A satisfying tradeoff seems to be found for $\alpha=0.2$ and $\beta=0.9$, which will be used in the results that follow.
        
        Both parameters determine the amount of transfer which occurs. However, they intervene in different ways. The parameter $\beta$ acts as a first rough pre-selection, and completely discards irrelevant sentences. Instead, $\alpha$ determines how strongly the queries from the reservoir are encouraged to be classified as a \textit{None} intent. Those which are classified as \textit{None} can nonetheless benefit the CVAE, as explained below.
        
        \begin{figure*}[htb]
            \centering
            \includegraphics[width=1.\textwidth]{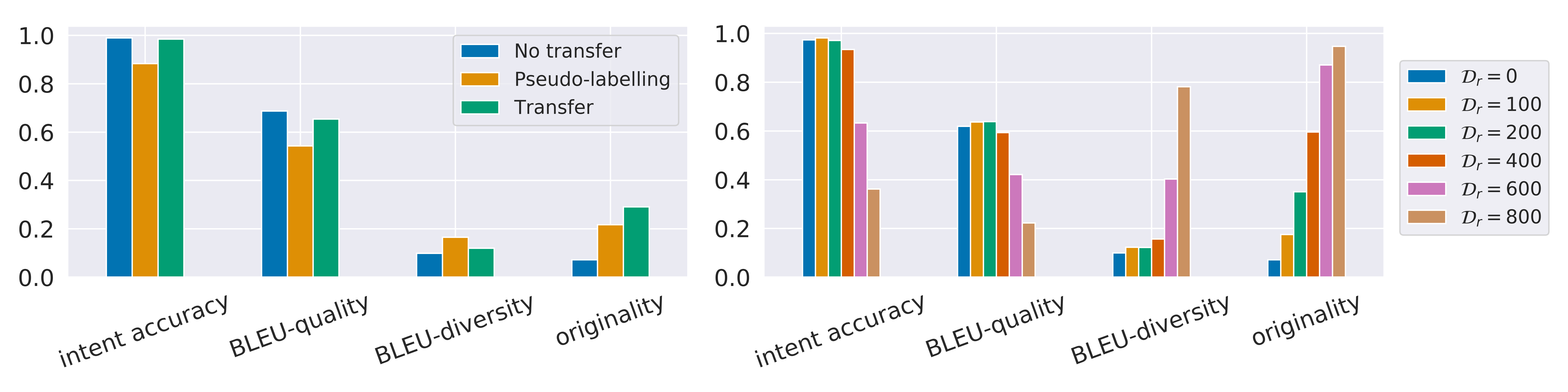}
            \caption{Generation metrics. (Left Panel) Comparison of our query transfer method with two baselines: one without any transfer and one with InferSent pseudo-labelling. (Right Panel) Effect of the size of the reservoir $\mathcal{D}_r$ on the generation metrics. We set $|{D}_0|=200$, $\alpha=0.2$, $\beta=0.9$}
            \label{fig:results}
         \end{figure*}
         
        \subsubsection{Efficiency of query transfer.}
        
        To test the efficiency of the query transfer, we compare it to two baselines. The first one is simply a CVAE trained only on $\mathcal{D}_0$ (in blue on the left panel of Fig.~\ref{fig:results}). The second one, referred to as \textit{pseudo-labelling} (in orange on the figure), leverages queries from $\mathcal{D}_r$ directly associated to intents of $\mathcal{D}_0$ using InferSent-based similarity scores (the CVAE is trained without a \textit{None} class and a sentence in $\mathcal{D}_r$ is mapped to its closest intent as defined in Section~\ref{sec:sent_selec}). For this second baseline, when the measured similarity exceeds a certain threshold for a given intent, the sentence from $\mathcal{D}_r$ is directly added to the corresponding intent in $\mathcal{D}_0$, on which the CVAE is trained. 
        
        The left panel of Fig.~\ref{fig:results} shows that the proposed query transfer method improves the diversity metrics (especially the originality) of the generated sentences, with hardly any deterioration in quality. In comparison, the pseudo-labelling approach significantly deteriorates  the quality of generated sentences. 
        This means that the proposed mechanism does more than extracting relevant queries from the reservoir: the underlying intuition is that the CVAE benefits from the irrelevant intents by improving its language modelling abilities, without corrupting the generation task thanks to the extra \textit{None} class.
    
        Finally, the right panel of Fig.~\ref{fig:results} 
        shows a remarkable improvement of the diversity metrics when the number of sentences injected from the reservoir $\mathcal{D}_r$ increases, without any loss in quality up to a certain point at which the quality degrades strongly. This is due to the important imbalance introduced in the conditioning mechanism of the CVAE, too much irrelevant content being passed onto the intent classes. A satisfying trade-off is found for $|\mathcal{D}_r|=|\mathcal{D}_0|$.
    
    \subsection{Results on data augmentation for language models}
        In this section, we show that our query transfer mechanism can effectively be used as data augmentation technique for language modeling tasks. Indeed, leveraging in-domain language models -- trained for a specific use case rather than in a large vocabulary setting -- allows to both reduce their size and increase their in-domain accuracy~\cite{SNIPS}. We compare the perplexity \cite{bahl1983maximum} of Language Models (LM) trained on three datasets: (i) $\mathcal{D}_0$, the initial dataset; (ii)  $\mathcal{D}_{\mathrm{aug}}$, containing $\mathcal{D}_0$ augmented by sentences generated by the CVAE model trained on $\mathcal{D}_0$ with query transfer; (iii) $\mathcal{D}_{\mathrm{ref}}$, containing  $\mathcal{D}_0$ augmented by ``real'' sentences from the original dataset as a reference point. 

        \subsubsection{Augmentation setup.}
            This experiment is done on the publicly-available Snips benchmark dataset \cite{coucke2018snips}. The different training sets $\mathcal{D}_0$ are drawn from the original set of 2000 queries per intent (files denoted \texttt{train\_IntentName\_full.json} in the dataset GitHub repository\footnote{\url{https://github.com/snipsco/nlu-benchmark/tree/master/2017-06-custom-intent-engines}}.
            The test set $\mathcal{D_{\mathrm{test}}}$ used to evaluate language model perplexities (see next section) consists in the validation sets of the Snips benchmark dataset (\texttt{validate\_IntentName.json} files).
            
            Four different data regimes are explored for training (i.e. $|\mathcal{D}_0|$ taking values in $[125, 250, 500, 1000]$). For each data regime, a CVAE is trained with 5 initialization seeds and a hyperparameter search on the embedding dimension and the encoder and decoder hidden sizes. In this experiment, we set $\alpha=0.2$, $\beta=0.9$ and $|\mathcal{D}_r|=|\mathcal{D}_0|$, consistently with results displayed in Section \ref{sec:results}. Three models are then selected based on the four metrics defined above, yielding 3 models per size of $\mathcal{D}_0$. Each model is used to generate a set of 1000 new queries used to augment the training set -- the newly formed training set being denoted $\mathcal{D}_{\mathrm{aug}}$. We consider two augmentation ratios: $+50\%$ ($|\mathcal{D}_{\mathrm{aug}}|=1.5\times |\mathcal{D}_{0}|$) and $+100\%$ ($|\mathcal{D}_{\mathrm{aug}}|=2\times |\mathcal{D}_{0}|$).
        
        \subsubsection{Perplexity computation.}
            The SRILM toolkit \cite{stolcke2002srilm} is used to train 4-grams language models with Kneser-Ney Smoothing \cite{kneser1995improved} on $\mathcal{D}_0$, $\mathcal{D}_{\mathrm{aug}}$, and $\mathcal{D}_{\mathrm{ref}}$ respectively, when varying the size of $\mathcal{D}_0$ and the augmentation ratio. Perplexities are only comparable if the vocabulary supported by the various models is the same. To fix this issue, the words contained in at least $\mathcal{D}_{0}$, $\mathcal{D}_{\mathrm{aug}}$ and $\mathcal{D}_{\mathrm{ref}}$ are added as unigrams with a count $1$ in every LM. The CVAE might generate sentences already present in $\mathcal{D}_{0}$ but every sentence is kept only once. The perplexity -- averaged over the 3 experiments -- is then evaluated on a pool $\mathcal{D_{\mathrm{test}}}$ of 700 test sentences.
        
        \subsubsection{Results.}
        
        \begin{table}
\centering
\caption{\label{tab:perplexity} Relative loss of perplexity (\%) with respect to LM trained on the original dataset $\mathcal{D}_0$, when varying the size of $\mathcal{D}_0$ and the augmentation ratio.
                Results can only be compared row-wise because of the vocabulary restriction}
                \vspace{0.2cm}
\begin{tabular}{|lclcllclcl|}
\hline
 &
  $|D_0|$ &
   &
  \begin{tabular}[c]{@{}c@{}} augmentation \\ ratio \end{tabular} &
   &
   &
 \begin{tabular}[c]{@{}c@{}}PPL(\%) \\  $\mathcal{D}_{\mathrm{aug}}$\end{tabular} &
   &
 \begin{tabular}[c]{@{}c@{}}PPL(\%) \\  $\mathcal{D}_{\mathrm{ref}}$\end{tabular}  &
   \\ \hline
 & \multirow{2}{*}{$125$}  &  & $+50\%$  &  &  & $-2.322$ &  & $-17.73$ &  \\
 &                         &  & $+100\%$ &  &  & $-5.909$ &  & $-28.62$ &  \\ \hline
 & \multirow{2}{*}{$250$}  &  & $+50\%$  &  &  & $-1.756$ &  & $-17.72$ &  \\
 &                         &  & $+100\%$ &  &  & $-3.755$ &  & $-22.85$ &  \\ \hline
 & \multirow{2}{*}{$500$}  &  & $+50\%$  &  &  & $-3.335$ &  & $-12.34$ &  \\
 &                         &  & $+100\%$ &  &  & $-4.046$ &  & $-18.55$ &  \\ \hline
 & \multirow{2}{*}{$1000$} &  & $+50\%$  &  &  & $-1.031$ &  & $-9.278$ &  \\
 &                         &  & $+100\%$ &  &  & $-0.511$ &  & $-13.62$ &  \\ \hline
\end{tabular}
\end{table}

            Table~\ref{tab:perplexity} shows the results when varying the size of $\mathcal{D}_{0}$ and the number of sentences generated by the augmentation process (augmentation ratios of $50\%$ and $100\%$). The perplexity computed on the test set of the Snips dataset is consistently lower when the LM is trained on $\mathcal{D}_{\mathrm{aug}}$ rather than 
            on $\mathcal{D}_{0}$ (though it does not reach the performance of augmentation with real data $\mathcal{D}_{\mathrm{ref}}$). The improvement is less significant as the dataset size increases, illustrating that most phrasings of the various intents are already covered in this data regime. These results are encouraging in the low data regime and we will evaluate our query transfer as a data augmentation process for SLU tasks in a  future work.

    \subsection{Examples of generated samples}
        In this section, we display examples of sentences generated with a CVAE trained on 250 queries (approximately 36 examples per intent). This trained model is used to generate 1000 new sentences, some of which are displayed in Table~\ref{tab:sentences} for the \textit{GetWeather} intent. The presented approach is able to generate patterns that were not present in the training set -- as underlined by the ``originality'' metrics -- but also patterns that are never seen in the full Snips benchmark dataset (14000 queries). Interestingly, the number of occurrences of patterns among all generated examples does not seem to be related to their prevalence in the training set, or even in the full Snips benchmark dataset. 
        
            

\begin{table}
            \centering
            \caption{\label{tab:sentences} Examples of generated delexicalized sentences (patterns) obtained with a CVAE trained on 250 examples, along with the corresponding occurrences among the 1000 generated sentences (leftmost column), the 250 training utterances (middle column) and the 14000 queries of the full Snips dataset (rightmost column). We used $\alpha=0.2$, $\beta=0.9$}
            \vspace{0.2cm}
            \begin{tabular}{|cccc|}
            \hline
            \# generated & \# in train & \# total  & generated pattern    \\ \hline
            
            13 & 1 & 8 & \begin{tabular}[c]{@{}c@{}} \small is it [ConditionDescription] in [City]\end{tabular} \\ 
                    
            21 & 1 & 13 & \begin{tabular}[c]{@{}c@{}} \small what is the weather forecast for [City]  \end{tabular} \\ 
            
            17 & 0 & 10 & \begin{tabular}[c]{@{}c@{}} \small what's the weather forecast for [City]\end{tabular} \\ 
            
            14 & 0 & 2 & \begin{tabular}[c]{@{}c@{}} \small tell me the weather forecast for [City] \end{tabular} \\
            
             8 & 0 & 0 & \begin{tabular}[c]{@{}c@{}} \small how's the weather supposed to be on [TimeRange] \end{tabular} \\ \hline
            
            \end{tabular}

        \end{table}

\section{Conclusion}
\label{sec: conclusions}
We introduce a transfer method to alleviate data scarcity in conditional generation tasks where one has access to a large unlabelled dataset containing some potentially useful information. We use conditional variational autoencoders with a transfer parameter $\alpha$ and a selectivity threshold $\beta$ which are both used to control the trade-off between quality and diversity of the data augmentation. We choose to focus on the low data regime, as it is the most relevant for customized closed-domain dialogue systems, where gathering manually annotated datasets is cumbersome.

    
    Transferring knowledge from the large reservoir dataset $\mathcal{D}_r$ to the original dataset $\mathcal{D}_0$ comes with the risk of introducing unwanted information which may corrupt the generative model. However, this risk may be controlled by adjusting two parameters. First, we consider a selectivity threshold $\beta$ to adjust how much irrelevant data is discarded from $\mathcal{D}_r$ during a pre-processing step. Second, we introduce a transfer parameter $\alpha$, adjusting the supervision of unlabelled examples from $\mathcal{D}_r$, low values of $\alpha$ facilitating transfer from the reservoir.
    
    
    While we assess the performance of the proposed generation technique by both introducing \textit{quality} and \textit{diversity} metrics and showing how the introduced parameters may help choosing the best trade-off, we also apply it as a data augmentation technique on a small language modelling task. The full potentiality of this method for more complex SLU tasks still needs to be explored and will be the subject of a future work.

%
%
\bibliographystyle{splncs04}
\bibliography{slsp2020}

\end{document}